\documentclass[review,12pt]{elsarticle}

\usepackage{amssymb}
\usepackage{amssymb}
\usepackage{times}  
\usepackage{helvet}  
\usepackage{courier}  
\usepackage{url}  
\usepackage{graphicx}  
\usepackage{dcolumn}
\usepackage{subfigure}
\usepackage{bm}
\usepackage{indentfirst}
\usepackage{amsmath,amsfonts,amssymb}
\usepackage[dvips]{epsfig}
\usepackage{epsfig}
\usepackage{lettrine}
\usepackage[dvips]{graphicx}
\usepackage{multirow}
\usepackage{color}
\usepackage{float}
\usepackage{algorithm}
\usepackage{algpseudocode}
\usepackage{multirow}

\DeclareMathOperator*{\argmin}{argmin}

\journal{Pattern Recognition}

\begin{document}

\begin{frontmatter}



\title{A Generalized Kernel Risk Sensitive Loss for Robust Two-Dimensional Singular Value Decomposition}


\author{Miaohua Zhang, Yongsheng Gao}

\address{School of Engineering and Built Environment, Griffith University, Australia.}

\begin{abstract}
Two-dimensional singular decomposition (2DSVD) has been widely used for image processing tasks, such as image reconstruction, classification, and clustering.  However, traditional 2DSVD algorithm is sensitive to outliers because it is based on the mean square error (MSE) loss which has the potential to magnify the reconstruction error. To overcome this problem, in this paper, we propose a robust 2DSVD framework based on a generalized kernel risk sensitive loss (GKRSL-2DSVD) which is more robust to noise and and outliers. Since the proposed objective function is non-convex, a majorization-minimization algorithm is developed to efficiently solve it with guaranteed convergence. The proposed framework has inherent properties of processing non-centered data, rotational invariant, being easily extended to higher order spaces. Experimental results on public databases demonstrate that the performance of the proposed method on different applications significantly outperforms that of all the benchmarks.
\end{abstract}


\end{frontmatter}


\section{Introduction}
\label{}
Subspace learning methods are usually considered for data dimensionality reduction to improve the efficiency of algorithms while keeping the energy of the data as much as possible. Principal component analysis (PCA)~\cite{turk1991eigenfaces}, as one of the most representative dimensionality reduction methods, aims to learn a projection matrix to project the high-dimensional data to a new space with lower dimensionality. Belhumeur et al.~\cite{belhumeur1997eigenfaces} proposed the linear discriminant analysis (LDA) to take the label information into consideration, and learn a projection matrix by maximizing the between-class variation while minimizing the within-class variation. However, both PCA and LDA are based on the Euclidean distance, and obtain optimal projections for linear data but inferior ones for nonlinear data, i.e., face image with illumination, expression, and pose changes. To solve this problem, He et al.~\cite{he2005face} argued that the real-world data resides in a low-dimensional manifold, and proposed a local preserving projections(LPP) algorithm to preserve the local structure of data.

However, the methods mentioned above are based on the mean square error (MSE) which is not a robust function because MSE-based loss function measures each representation error point equally, which will cause biased solution when there are large noise and outliers. In past decades, non-second order based subspace algorithms have been verified to have advantages in reducing the influence of outliers in the data. For example, Ke and Kanade~\cite{ke2005robust} measures the representation error using the $L_1$-norm and proposed the $L_1$-PCA. Kwak~\cite{kwak2008principal} proposed the greedy $L_1$-PCA, and Nie et al.~\cite{nie2011robust} improve the efficiency and performance of the $L_1$-PCA with a non-greedy style. $L_1$-norm based LDA and LPP methods has also been extensively studied in past years, such as $L_1$-norm LDA~\cite{zhong2013linear} and $L_1$-norm LPP~\cite{zhong2014discriminant}. Since the $L_1$-norm PCA is not rotational invariant, to solve this problem, Ding et al.,~\cite{ding2006r} proposed an $R_1$-PCA by taking advantages of the robustness of $L_1$-norm and rotational invariant property of $L_2$-norm. There are also some subspace learning methods based on robust metrics. For example, He et al.~\cite{he2011robust} proposed a HQ-PCA based on the maximum correntropy criterion (MCC) with discriminative weights for normal clean data and outliers.

  However, the methods mentioned above are all based on the vector space, which will destroy the inherent structure of an image since the real-world data is always captured with multiple dimensions. To better exploit the spatial information carried by image, Yang et al.~\cite{yang2004two} proposed two-dimensional PCA algorithm (2DPCA) and 2DLDA~\cite{yang2005two} which directly apply PCA method to 2D images. To improve the robustness of these 2D methods against outliers, Li et al.~\cite{li2010l1} take the advantages of $L_1$-norm, and proposed $L_1$-2DPCA. Wang et al.~\cite{wang2017two} find that the 2DPCA is based on a squared F-norm which may yield suboptimal solution when there are outliers. To solve this problem, they developed a robust 2DPCA based on the F-norm.

  Unlike 2DPCA that employs a one-sided transformation, Ye~\cite{ye2005generalized} proposed a two-sided linear transformation called the generalized low-rank approximations of matrices (GLRAM) and used an iterative procedure to solve it. Ding and Ye~\cite{ding20052} proposed a non-iterative two dimensional singular value decomposition (2DSVD) algorithm. Huang and Ding~\cite{huang2008robust} took the rotational invariance property of the $R_1$-norm and applied it to 2DSVD and higher tensor factorization. Although the outlier resistant ability of the above methods has been much improved, they treat each training sample equally without any discriminative constraints for inliers (normal data) and outliers. Motivated by the successful of the information theoretic learning (ITL) based criterions in enhancing the robustness of learning algorithms~\cite{he2011robust, chen2018mixture,wu2019correntropy,liangjun2018correntropy}, we propose a generalized kernel risk sensitive loss (GKRSL) to overcome the vulnerability of 2DSVD in dealing with outliers. The KRSL~\cite{chen2017kernel,rastegarnia2018tracking,xing2018robust,zhang2019kernel} is inspired by the risk sensitive loss and MCC whose surface is highly non-convex, i.e., the surface far away from the optimal solution is flat while the area around the optimal solution is very steep, which may yield a suboptimal solution. KRSL not only improves the convexity of MCC but also remains its outlier-resistance ability. However, the KRSL is a kernel based similarity measurement defined in a second order space. The second order measurement may not always the best choice in matching the representation error. The proposed GKRSL offers more flexibility in controlling the error, thus a more robust solution is obtained. The contributions of this paper are summarized as follows:
 \begin{itemize}
  \item A generalized kernel risk sensitive loss (GKRSL) is first defined in this paper, and a robust GKRSL based 2DSVD algorithm is proposed.
 \item A new Majorization Minimization optimization procedure is developed to solve the GKRSL-2DSVD with guaranteed convergence.
 \item An extension of the proposed algorithm to higher order space is provided.
 \item The GKRSL-2DSVD algorithm is rotational invariant, and the data mean can be automatically updated during the optimization to exclude the information of outliers.
 \end{itemize}

\section{Related Works}
 Denoted by $\{X_1,X_2,\ldots,X_N\}$ a set of matrix-based samples, and each sample $X_i$ is an image with size $m\times n$. Ding and Ye~\cite{ding20052} directly applied matrix decomposition on 2D images and proposed the 2DSVD method which computes a two-sided low-rank approximation of matrices by minimizing an approximation error as follows:
 \begin{equation}
\underset{L,M, R}{\min} J(L,M,R)=\sum_{i=1}^N\|\hat{X}_i-LM_iR^T\|_F^2,
 \end{equation}
 where $\hat{X}_i=(X_i -\bar{X})$ denotes the data after subtracting $\bar{X}$ from $X_i$, and $\bar{X}=\frac{1}{N}\sum_{i=1}^N X_i$ is the mean image of the dataset. $L\in\Re^{m\times k_1}$, $M=\{M_i\}_{i=1}^N$, $R\in\Re^{n\times k_2}$, and $M_i\in\Re^{k_1\times k_2}$. The covariance matrices are calculated from two directions, i.e., row-row and column-column, as
 \begin{equation}
 C_1=\sum_{i=1}^N \hat{X}_iRR^T\hat{X}_i^T, ~~~
 C_2=\sum_{i=1}^N \hat{X}_i^TLL^T\hat{X}_i.
\end{equation}

The optimal $L$ and $R$ can be obtained by calculating the largest $k_1$ and $k_2$ eigenvectors of $C_1$ and $C_2$, respectively.

From (1), we know that the 2DSVD adopts the $L_2$ norm based objective function, which is likely to magnify the effect of heavy noise or outliers. Huang and Ding~\cite{huang2008robust} talked about using an $L_1$ norm based cost function to overcome this drawback. However, $L_1$ norm based 2DSVD algorithm is computational expensive and is rotational variant. Then they proposed a rotational invariant 2DSVD ($R_1$-2DSVD) algorithm by taking advantages of the outlier resistance ability of the $L_1$ norm and the rotational invariance property of the $L_2$ norm. The objective function of $R_1$-norm based 2DSVD is defined as:
 \begin{equation}
 \underset{L, M, R}{\min} J(L,M,R)=\sum_{i=1}^{N} \sqrt{\|\hat{X}_i-LM_iR^T\|^2},
\end{equation}
where $L$, $M$, and $R$ are the same size as that defined in (1).
Different from the original 2DSVD in (1), the optimal projection matrices $L$ and $R$ in $R_1$-2DSVD can be obtained by calculating the eigenvectors of the following two reweighted covariance matrices $C_1$ and $C_2$:
 \begin{equation}
 C_1=\sum_{i=1}^N w_i~\hat{X}_iRR^T\hat{X}_i^T ,~~~C_2=\sum_{i=1}^N w_i~\hat{X}_i^TLL^T\hat{X}_i,
 \end{equation}
where $w_i=1/\sqrt{\textrm{Tr}(\hat{X}_i^T\hat{X}_i-\hat{X}_i^TLL^T\hat{X}_iRR^T )}$.

However, 2DSVD directly decompose the data matrix without discriminative constraints for outliers which will skew the learned features. Although the R1-2DSVD is more robust than 2DSVD, it also treats each sample equally. Moreover, 2DSVD and $R_1$-2DSVD assume that the training samples are centered data, which is not difficult to ensure in actual applications. Thus, to solve these problems in one single model, we propose a more robust GKRSL-2DSVD algorithm in the following sections.

\section{Proposed Method}

 \subsection{Definition of the generalized kernel risk sensitive loss (GKRSL)}
 Denote by $A$ and $B$ the two random variables, the GKRSL is given by
 \begin{equation}
\begin{aligned}
&f_{\text{GKRSL}}(A-B)=\frac{1}{\lambda}\mathbf{E}\left[\text{exp}\left(\lambda\eta\|\kappa(A)-\kappa(B)\|_{H}^{p}\right)\right]\\
&=\frac{1}{\lambda}\mathbf{E}\left[\text{exp}\left(\lambda\eta\left(\|\kappa(A)-\kappa(B)\|_{H}^2\right)^{\frac{p}{2}}\right)\right]\\
&=\frac{1}{\lambda}\mathbf{E}[\text{exp}(\lambda(1-g_{\sigma}(A-B))^{\frac{p}{2}})],
 \end{aligned}
 \end{equation}
where $\eta=2^{-\frac{p}{2}}$, $p>0$ is the order of error loss~\cite{chen2017kernel,xing2018robust}, and GKRSL reduces to KRSL when $p$ is 2.  $\lambda>0$ is a parameter that controls the convexity of the function, $\mathbf{E}(x)$ is the expectation of $x$. $g_{\sigma}(x)$ is a Mercer kernel with the bandwidth being $\sigma$, $\kappa(x)$ is a nonlinear mapping function that maps the variable $x$ from the original space to the kernel spaces, thus (5) can also be regarded as a similarity measurement between $A$ and $B$ in the kernel space $H$. In actual implementation, only a finite number of samples are available, the Parzen windowing method can be applied to estimate the GKRSL on a finite number of available samples $\{(a_i,b_i)\}_{i=1}^N$ \cite{liu2007correntropy,he2011maximum}:
\begin{equation}
f_{\text{GKRSL}}(A-B)=\frac{1}{N\lambda}\sum_{i=1}^N\text{exp}(\lambda(1-k_{\sigma}(a_i-b_i))^{\frac{p}{2}}).
\end{equation}
The GKRSL model actually measures the distance between $A=[a_i,a_2,\cdots,a_N]^T$ and $B=[b_1,b_2,\cdots,b_N]^T$.
\subsection{The Proposed GKRSL-2DSVD}
Compared with MCC and KRSL, the GKRSL gives a more flexible choice in controlling the representation error, and thus the error fitting ability will be much enhanced.
Motivated by the advantages of the GKRSL model in modeling the error, we propose the following GKRSL model to learn more robust features for 2DSVD in the presence of outliers.

 \begin{equation}
 \begin{aligned}
&\underset{L,R,\{M_i\},\bar{X}}{\min}~ f_{\text{GKRSL}}(E_i)\\
&=\frac{1}{N\lambda}\sum_{i=1}^N \text{exp}(\lambda(1-\text{exp}(-\frac{E_i^2}{2\sigma^2}))^{\frac{p}{2}})\\
&\text{subject to (s.t.)}~~~ L^TL=I,~~~ R^TR=I,\\ &~~~~~~~~~~~~~~~~~~~~~~~~~~E_i=\sqrt[]{\|\hat{X}_i-LM_iR^T\|_F^2}.
\end{aligned}
 \end{equation}

To solve the problem in (7), we first calculate $M_i$ by setting the derivative of $f_{\text{GKRSL}}$ with respect to $M_i$ to zero:
\begin{equation}
\begin{aligned}
&\frac{\partial f_{\text{GKRSL}}}{\partial M_i}=\frac{p\lambda}{2\sigma_2}Q_1Q_2Q_3(X_i-LM_iR^T)L^TR=0.\\
&~~~~~~~~~~Q_1=\text{exp}(\lambda(1-\text{exp}(-\frac{E_i^2}{2\sigma^2}))^{\frac{p}{2}}),\\
&~~~~~~~~~~Q_2=(1-\text{exp}(-\frac{E_i^2}{2\sigma^2}))^{\frac{p}{2}-1},\\
&~~~~~~~~~~Q_3=\text{exp}(-\frac{E_i^2}{2\sigma^2}).
\end{aligned}
\end{equation}

Since $Q_1$,$Q_2$, and $Q_3$ are not possible to be zeros, the term $X_i-LM_iR^T$ should be zero, then we have $M_i=L^TX_iR$. Then by applying $M_i$ to the objective function (7), we have
\begin{equation}
 \begin{aligned}
&\underset{L,R,\bar{X}}{\min}~f_{\text{GKRSL}}(E_i)\\
& =\frac{1}{N\lambda}\sum_{i=1}^N \text{exp}(\lambda(1-\text{exp}(-\frac{E_i^2}{2\sigma^2}))^{\frac{p}{2}})\\
& ~~~~~\text{s.t.}~~ L^TL=I, R^TR=I,\\
& ~~~~~E_i=\sqrt{\|\hat{X}_i-LL^T\hat{X}_iRR^T\|_F^2}.
\end{aligned}
 \end{equation}
\section{Optimization by Majorization Minimization}
Due to that the function in (9) is non-convex, in this section, we introduce how to solve this non-convex optimization problem via a majorization minimization (MM) algorithm. In MM, instead of solving the complicated non-convex optimization problem directly, it replaces the original function with its upper-bound surrogate function in the majorization step and then minimize the resulted function in the minimization step.
The MM algorithm works by repeating the following two steps\\
(1) construct a convex upper bound function of the non-convex objective function, i.e., $f_{\text{GKRSL}}(E|E_t)$.\\
(2) minimize the surrogate function $f_{\text{GKRSL}}(E|E_t)$ until convergence.
\subsection{Majorization Procedure}
We here introduce how to construct the surrogate function. Since the function $f_{\text{GKRSL}}(E)$ is non-decrease and non-convex, the first Taylor expansion of $f_{\text{GKRSL}}(E)$ in the proximity point $E_{i,t}$ satisfies
\begin{equation}
\begin{aligned}
&f_{\text{GKRSL}}(E_i)\leq f_{\text{GKRSL}}(E_{i,t})+f'(E_{i,t})(E_i-E_{i,t})+c\\
&~~~~~~~~~~~~~~~~~~~=f_{\text{GKRSL}}(E_i|E_{i,t}),
\end{aligned}
\end{equation}
where $c$ is a constant, and $t$ is the iteration number.

According to the MM theory in~\cite{hunter2004tutorial}, we have
\begin{equation}
f(E)\leq f_{\text{GKRSL}}(E|E_t), ~~~\text{and}~~~f_{\text{GKRSL}}(E_t)=f_{\text{GKRSL}}(E_t|E_t).
\end{equation}

If the $E_{t+1}$ denotes the minimizer of the  $f_{\text{GKRSL}}(E|E_t)$, then the MM procedure has the descent property as
\begin{equation}
f_{\text{GKRSL}}(E_{t+1})\leq f_{\text{GKRSL}}(E_{t}), ~~~~~~t=1,2,\cdots.
\end{equation}

Then the objective function can be upperbounded by $f'_{\text{GKRSL}}(E_{t})E$ by omitting the constant terms in $f_{\text{GKRSL}}(E|E_{t})$, thus we have
\begin{equation}
\min f_{\text{GKRSL}}(E) \leq f'_{\text{GKRSL}}(E_t)E.
\end{equation}

\subsection{Minimization Procedure}
Based on the above analysis, minimizing (9) can be achieved by minimizing the following surrogate function
\begin{equation}
\begin{aligned}
& \underset{L,R,\bar{X}}{\argmin}~~~f_{\text{GKRSL}}(E|E_t) \\
& \text{s.t.}~~~ L^TL=I,~~~ R^TR=I, E_i=\sqrt{\|\hat{X}_i-LL^T\hat{X}_iRR^T\|_F^2}.
\end{aligned}
\end{equation}
 The Lagrangian function of (14) is
 \begin{equation}
 \begin{aligned}
 &\mathcal{L}(\hat{L},\hat{R},\hat{\bar{X}})\\
 &=f_{\text{GKRSL}}(E|E_t)+\text{Tr}(\Omega_1(L^TL-I))+\text{Tr}(\Omega_2(R^TR-I)),
 \end{aligned}
 \end{equation}
where $\text{Tr}(x)$ is the trace of $x$. According to (10), we have
\begin{equation}
\begin{aligned}
& f_{\text{GKRSL}}(E|E_t)=f_{\text{GKRSL}}'(E_t)E =\frac{p}{2} P_1P_2P_3E_t E\\
& ~~~~~~~~P_1=\text{exp}(\lambda(1-\text{exp}(-\frac{E_t^2}{2\sigma^2}))^{\frac{p}{2}}),\\
&~~~~~~~~P_2=(1-\text{exp}(-\frac{E_t^2}{2\sigma^2}))^{\frac{p}{2}-1},\\
&~~~~~~~~P_3=\text{exp}(-\frac{E_t^2}{2\sigma^2}).
\end{aligned}
\end{equation}
Let $W=\frac{p}{2}P_1P_2P_3E_t$ be the weight for each sample. Thus (16) can be rewritten as
\begin{equation}
\begin{aligned}
&\underset{\hat{L},\hat{R},\hat{\bar{X}}}{\argmin}~~\mathcal{L}\{\hat{L},\hat{R},\hat{\bar{X}}\}\\
&=\frac{1}{N}\sum_{i=1}^N W_iE_i+\text{Tr}(\Omega_1(L^TL-I))+\text{Tr}(\Omega_2(R^TR-I)).\\
&=\frac{1}{N}\sum_{i=1}^N W_i\sqrt{\|\hat{X}_i-LL^T\hat{X}_iRR^T\|_F^2}\\
&~~~~~~~~~~+\text{Tr}(\Omega_1(L^TL-I))+\text{Tr}(\Omega_2(R^TR-I))\\
&=\frac{1}{N}\sum_{i=1}^N W_i\sqrt{\text{Tr}(\hat{X}_i^T\hat{X}_i-\hat{X}_i^TLL^T\hat{X}_iRR^T)}\\
&~~~~~~~~~~+\text{Tr}(\Omega_1(L^TL-I))+\text{Tr}(\Omega_2(R^TR-I)),\\
&\text{s.t.} ~~~W_i=\frac{p}{2}P_1P_2P_3 E_{i,t}, ~~~~\hat{X}_i=X-\bar{X}.
\end{aligned}
\end{equation}

The optimal solution $\{\hat{L},\hat{R},\hat{\bar{X}}\}$ can be obtained by setting the derivative of Lagrangian function in (17) with respect to (w.r.t.) $L$, and $R$, $\hat{\bar{X}}$, respectively.
First, by solving the following problem, we can obtain the optimal mean matrix $\hat{\bar{X}}$.
\begin{equation}
\begin{aligned}
&\frac{\partial \mathcal{L}}{\partial \bar{X}}=\frac{\partial \sum_{i=1}^N W_i\sqrt{\|\hat{X}_i-LL^T\hat{X}_iRR^T\|_F^2}}{\partial \bar{X}}\\
&=\frac{\partial \sum_{i=1}^N W_i\sqrt{\|X_i-\bar{X}-LL^T(X_i-\bar{X})RR^T\|_F^2}}{\partial \bar{X}}=0.
\end{aligned}
\end{equation}
 By solving (18), the optimal $\bar{X}$ can obtained by
 \begin{equation}
 \bar{X}=\sum_{i=1}^N\frac{ \frac{1}{2}W_iX_i}{\sqrt{\text{Tr}(\hat{X}_i^T\hat{X}_i-\hat{X}_i^TLL^T\hat{X}_iRR^T)}}/\sum_{i=1}^N W_i.
 \end{equation}

 We solve the optimal $\hat{L}$ by setting the derivative of Lagrangian function w.r.t. $L$ as follows.
 \begin{equation}
 \begin{aligned}
 &\frac{\partial \mathcal{L}}{\partial L}=\frac{-W_i\hat{X}_iRR^T\hat{X}_i^TL}{2\sqrt{\text{Tr}(\hat{X}_i^T\hat{X}_i-\hat{X}_i^TLL^T\hat{X}_iRR^T)}}+\Omega_1L=0\\
 &~~~~~~=-FL+\Omega_1L=0\\
 &\Longrightarrow ~~~~~~FL=\Omega_1L.
 \end{aligned}
 \end{equation}

 The optimal $\hat{L}$ can be updated by the largest $k_1$ eigenvectors of covariance matrix $F$.

 Then, we solve the optimal $\hat{R}$ by setting the derivative of Lagrangian function w.r.t. $R$ as follows.
\begin{equation}
\begin{aligned}
&\frac{\partial \mathcal{L}}{\partial R}=\frac{-W_i\hat{X}_i^TLL^T\hat{X}_iR}{2\sqrt{\text{Tr}(\hat{X}_i^T\hat{X}_i-\hat{X}_i^TLL^T\hat{X}_iRR^T)}}+\Omega_2R=0\\
 &~~~~~~=-GR+\Omega_2R=0\\
 &\Longrightarrow ~~~~~~GR=\Omega_2R.
\end{aligned}
\end{equation}

The optimal $\hat{R}$ can be updated by the largest $k_2$ eigenvectors of covariance matrix $G$. This alternate optimization procedure is repeated until the error between the current $L$ and $R$ and the ones calculated in the last iteration falls bellow a prescribed threshold $\epsilon$.

Based on the above analysis, the robust GKRSL-2DSVD algorithm is summarized in Algorithm 1.
\begin{algorithm}[htb]
\caption{GKRSL-2DSVD  Algorithm}
\begin{algorithmic}[1]
\Require
   Given a data matrix $X=\{X_1,X_2,\cdots, X_N\}$ with each $X_i\in R^{m\times n}$. $p$, $\lambda$, $k_1$, $k_2$, and threshold $\epsilon$.
\Ensure
 $\{W_i\}_{i=1}^N$, $\hat{L}, \hat{R}, \hat{\bar{X}}$.
 \While {$t$=1,\dots,$T$}
\State Update $\{W_i\}_{i=1}^N$ for each sample  by (16).
\State Update $\hat{\bar{X}}$ by (19).
\State Update $L$ and $R$ by (20) and (21), respectively.
\State 1) Update the covariance matrix $F$ with the latest projection matrices $L$ and $R$, the optimal $\hat{L}$ can be obtained by calculating the largest $k_1$ eigenvectors of $F$,
\State 2) Update the covariance matrix $G$ with the latest projection matrices $L$ and $R$, the optimal $\hat{R}$ can be obtained by calculating the largest $k_2$ eigenvectors of $G$,
\If {$\epsilon>1e-5$}
\State repeat;
\Else
\State $t\leftarrow t+1$; Break;
\EndIf
\EndWhile
\end{algorithmic}
\end{algorithm}

\subsection{Higher Order Extension}
 To process the image data with higher dimensional structure, we then extend the proposed algorithm to higher order spaces. Assume there is an $N$-dimensional data $\mathcal{X}=\{\mathcal{X}_{i_1i_2\cdots i_N}\}$ with  $i_1=1,\cdots,N_1$; $i_2=1,\cdots, N_2;\cdots; i_N=1,\cdots,N_n$. $\mathcal{X}$ can also be viewed as a set of tensor data $\{\mathcal{X}_{N_1},\mathcal{X}_{N_2},\cdots,\mathcal{X}_{N_n}\}$ where each $\mathcal{X}_i$
is an $(N-1)$-dimensional tensor. We compress ($N$-1) dimensions of each tensor $\mathcal{X}_{N_i}$  but not on the data index dimension~\cite{ding2006r}. The robust version of $N$-1 tensor decomposition based on the  proposed GKRSL is:
\begin{equation}
\begin{aligned}
&\underset{V_1,V_2,\cdots,V_N,\bar{\mathcal{X}}}{f_{\text{GKRSL}}}(\mathcal{E}|\mathcal{E}_t)\\
& \text{s.t.}~~~~~ V_n^TV_n=I,~~~~~n=1,\cdots, N-1.\\
& \mathcal{E}_i=\sqrt{\|\hat{\mathcal{X}}_{i_N}-V_1\otimes_1V_2\cdots V_{N-1}\mathcal{M}_{i_N}\|^2},
\end{aligned}
\end{equation}
where $\hat{\mathcal{X}}_{i_N}=\mathcal{X}_{i_N}-\bar{X}_{i_N}$, $\bar{X}_{i_N}=\frac{1}{N_n}\sum_{i_N=1}^{N_n}\mathcal{X}_{i_N}$, and $V_n\otimes_n\mathcal{M}$ denotes the $n$-mode tensor product of matrix $V_n$ and core tensor $\mathcal{M}$.
\begin{equation}
\begin{aligned}
& f_{\text{GKRSL}}(\mathcal{E}|\mathcal{E}_t)=f_{\text{GKRSL}}'(\mathcal{E}_t)\mathcal{E} =\frac{p}{2} A_1A_2A_3E_t E= W\mathcal{E}.
\end{aligned}
\end{equation}
where $A_1=\text{exp}(\lambda(1-\text{exp}(-\frac{\mathcal{E}_t^2}{2\sigma^2}))^{\frac{p}{2}})$, $A_2=(1-\text{exp}(-\frac{\mathcal{E}_t^2}{2\sigma^2}))^{\frac{p}{2}-1}$,and $A_3=\text{exp}(-\frac{\mathcal{E}_t^2}{2\sigma^2})$, $\mathcal{E}_t=\sqrt{\|\hat{\mathcal{X}}_{i_N}-V_1\otimes_1V_2\cdots V_{N-1}\mathcal{M}_{i_N}\|^2}$. Then the Lagrangian function for (22) is given by
\begin{equation}
\begin{aligned}
\mathcal{L}(\bar{\mathcal{X}},\{V_n\},\mathcal{M}_{i_N})=\frac{1}{N}\sum_{i_N=1}^N W_i\mathcal{E}_i+\sum_{i=1}^N \text{Tr}(\Omega_n(V_n^TV_n-I)),
\end{aligned}
\end{equation}
where $\{\Omega_n\}_{n=1}^{N-1}$ are symmetric Lagrangian Multipliers, the derivative of $\mathcal{L}$ w.r.t. the optimal solutions must be zeros. Then the optimal solution $\hat{\bar{X}}$ can be updated by
\begin{equation}
\begin{aligned}
\hat{\bar{\mathcal{X}}}=\sum_{i_N=1}^N \frac{\frac{1}{2}W_i \hat{\mathcal{X}}_i}{\sqrt{\|\hat{\mathcal{X}}_{i_N}-V_1\otimes_1V_2\cdots V_{N-1}\mathcal{M}_{i_N}\|^2}}/ \sum_{i=1}^N W_i.
\end{aligned}
\end{equation}
Then the projection matrices $\{V_n\}_{n=1}^{N-1}$ can be updated by \begin{equation}
\begin{aligned}
&\frac{\partial \mathcal{L}}{\partial V_n}=\frac{-W_i\sum_{i_{-n}}(\hat{\mathcal{X}}_{i_1,\cdots,i_{N-1}}^{i_N}\hat{\mathcal{X}}_{i_1^{'},\cdots,i_{N-1}^{'}}^{i_N} Z_{-n})V_n}{\sqrt{\|\hat{\mathcal{X}}_{i_N}-V_1\otimes_1V_2\cdots V_{N-1}\mathcal{M}_{i_N}\|^2}}\\
&~~~~~~~~~~~~~+2\Omega_n V_n=0,\\
&\Rightarrow ~~~~~~~ HV_n=\Omega_n V_n,
\end{aligned}
\end{equation}
where  $Z_{-n}=(V_1V_1^T)_{i_1 i_1^{'}} \dots (V_{n-1}V_{n-1}^T)_{i_{n-1}i_{n-1}^{'}}(V_{n+1}\\
V_{n+1}^T)_{i_{n+1}i_{n+1}^{'}}$, $W_i=\frac{p}{2}A_1A_2A_3E_t$, $i_{-n}=i_1i_1^{'},\ldots,i_{n-1}\\
i_{n-1}^{'},i_{n
+1}i_{n+1}^{'},\ldots,i_{N-1}i_{N-1}^{'}$, and $i_ni_n^{'}$ denotes the index of matrix $V_n$. Thus $V_n$ can be obtained by solving the eigenvectors of $H$.
\subsection{Convergence Analysis}
The convergence analysis of MM has been well studied by~\cite{hunter2004tutorial,sun2016majorization}, now we give the convergence analysis of GKRSL-2DSVD by the following theorems. According to the theory of MM and (11), we have
\begin{equation}
f_{\text{GKRSL}}(E^{t+1})\leq f_{\text{GKRSL}}(E_t),
\end{equation}
which indicates that the cost function in (7) is monotonically decreasing by using the proposed method in Algorithm 1  .

Since the the optimal $L$ and $R$ are updated alternatively, we then analyze the convergence of the objective function with respect $L$ and $R$, respectively.

\textbf{Theorem}: Algorithm 1 has a converged solution $L^*$ for problem (7), then $L^*$ satisfies the Karush-Kuhn-Tucker (KKT) condition of problem (7) with the constraint $(L^*)^TL^*=I$.

\textbf{Proof:} The Lagrangian function of objective function (7) w.r.t. the constraints $L^TL=I$ is given by
\begin{equation}
\begin{aligned}
\mathcal{L}_1=\frac{1}{N\lambda}\sum_{i=1}^N~ \text{exp} (\lambda(1-\text{exp}(-\frac{E_i^2}{2\sigma^2}))^{\frac{p}{2}})+\Omega_1(L^TL-I).
\end{aligned}
\end{equation}

According to the KKT condition of the optimization problem in (7), the derivative of $\mathcal{L}_1$ w.r.t. $L$ must be zero, which can be written as,
\begin{equation}
\begin{aligned}
&\frac{\partial\mathcal{L}_1}{\partial L}=-\frac{1}{N}\sum_{i=1}^N \frac{p}{2} O_1O_2O_3 X_iRR^TX_iL+2\Omega_1L=0\\
\end{aligned}
\end{equation}
By using some simple algebra operations, (29) can be rewritten as
\begin{equation}
\frac{1}{N}\sum_{i=1}^N \frac{p}{2} O_1O_2O_3 X_iRR^TX_iL=\Omega_1L.
\end{equation}

According to Algorithm 1, the optimal solution $L$ can be found by solving the cost function (14). Thus Algorithm 1 has a converged solution that satisfies the KKT condition of the cost function (14). The Lagrangian function of (14) is
\begin{equation}
\begin{aligned}
&\mathcal{L}_2=\frac{1}{N}\sum_{i=1}^N W_i \sqrt{\text{Tr}(\hat{X}_i^T\hat{X}_i-\hat{X}_i^TLL^T\hat{X}_iRR^T)}\\
&~~~~~~~~~~~~~~+\Omega_1(L^TL-I).
\end{aligned}
\end{equation}

According to the KKT condition of the problem (14), the derivative of $\mathcal{L}_2$ w.r.t. $L$ must be zero:
\begin{equation}
\begin{aligned}
&\frac{\partial\mathcal{L}_2}{\partial L}=\frac{1}{N}\sum_{i=1}^N -W_i \frac{X_iRR^TX_i^TL}{\sqrt{\text{Tr}(\hat{X}_i^T\hat{X}_i-\hat{X}_i^TLL^T\hat{X}_iRR^T)}}\\
& ~~~~~~~~~~~~+\Omega_1L=0.
\end{aligned}
\end{equation}

Considering that $W=\frac{p}{2}P_1P_2P_3E_t$, and by using some simple algebra operations, we have
\begin{equation}
\begin{aligned}
\frac{1}{N}\sum_{i=1}^N \frac{p}{2} P_1P_2P_3 X_iRR^TX_iL=\Omega_1L.
\end{aligned}
\end{equation}

The term $P_1P_2P_3X_iRR^TX_i$ closely relates to $L_t$, assume that the local solution in $(t+1)$th iteration is $L^*$, we have $L^*=L^{t+1}=L^t$. Since $P_1,P_2$ and $P_3$ have the similar form with $O_1,O_2$, and $O_3$, thus, in this case, (33) is just the same as (30). It means that Algorithm 1 has a converged solution that satisfies the KKT condition of (7), then we have
\begin{equation}
\frac{\partial \mathcal{L}}{\partial L}\bigm|_{L=L^*}=0.
\end{equation}

Based on the above analysis, we can say that the converged solution $L$ of Algorithm 1 is exactly a local solution of (7). For the optimal solution $R$, we also have the similar convergence analysis above.

 \section{Experimental results}
 \subsection{Databases and Parameter Settings}
 To verify the effectiveness of the proposed algorithm, in this section, we carry out extensive experiments on three public databases, including MNIST\footnotemark\footnotetext[1]{http://yann.lecun.com/exdb/mnist/} Handwritten Digit Database, ORL Face Database\footnotemark\footnotetext[2]{http://www.cl.cam.ac.uk/research/dtg/attarchive/facedatabase.html}, and YALE Face Database\footnotemark\footnotetext[3]{http://cvc.cs.yale.edu/cvc/projects/yalefaces/yalefaces.html}, for three different image processing applications, such as image classification, clustering, and reconstruction. The proposed algorithm is tested via different evaluation measurements and compared with seven classical two-dimensional subspace learing algorithms, including 2DPCA~\cite{yang2004two}, $L_1$-2DPCA~\cite{li2010l1}, F-2DPCA~\cite{wang2017two}, 2DSVD~\cite{ding20052}, $R_1$-2DSVD~\cite{huang2008robust}, N-2DNPP~\cite{zhang2017robust}, and S-2DNPP~\cite{zhang2017robust}.

$\lambda$ and $p$ are two important parameters in the proposed GKRSL-2DSVD algorithm where $\lambda$ controls the convex range, and $p$ controls the representation error distribution. In this paper $\lambda$ and $p$ are empirically set to  $\lambda=8$ and $p=8$ for experiments of image classification and clustering, and $\lambda=0.5$ and $p=0.5$ for image reconstruction. All the experiments are conducted on MATLAB R2015a.

\subsection{Experiments for Image Classification}
In this experiment, the MNIST handwritten digit database are used to test our algorithms in the presence of outliers. There are 60,000 training samples and 10,000 testing samples in this database, and all the digit images have been centered in a fixed-size of $28\times 28$. We respectively choose $\{200, 400, 600, 800\}$ samples from each digit (class) from the training set for training, and use all the testing samples for testing. All the samples are normalized by their norms, i.e., $X/\text{norm}(X)$. To simulate outliers, we random choose $5\%$ of the training samples and weight them by a magnitude $a$. i.e., $X_{o}=aX_c$ where $X_{o}$ and $X_c$ denote the simulated outlier image and clean image. We first set the magnitude of the outliers to $50$ ($a=50$) and number of principal components to $k_1=k_2=15$ to evaluate the performance of the proposed method under varying number of training samples in the presence of outliers. 1 nearest neighbor (1NN) is used as the classifier for all the algorithms. The classification accuracies of different algorithms using above settings are listed in Table 1 with the best results marked in bold. All the results are reported over 20 random trials to reduce the statistical deviation. The results in this table show that the recognition rates from different algorithms increase with the increase of the number of the training samples. The proposed algorithm outperforms all the benchmark methods under different size of training samples.

To check the influence of varying magnitude of outliers on the classification accuracy, we test all the algorithms on the $400\times 10$ training samples, and make $a$ vary from 20 to 100. Other parameters are the same as that used in Table 1. The classification accuracies are shown in Figure 1 (a), from which we can see that the performance of the proposed algorithm are almost unaffected under different $a$ while the accuracies of other algorithms reduces rapidly when $a$ increases.

To have a more intuitive analysis about the effect of different $\lambda$ and $p$ on the performance of the proposed algorithm, we visually display the classification accuracy under different $\lambda$ and $p$ using a bar chart in Figure 1(b). We can see that, with a fixed $p$ value, the accuracy increases with the $\lambda$ increasing. When the $\lambda$ is set to a fixed value, the accuracies increase fast with the increase of $p$.

\begin{table*}  
\caption {The recognition accuracy of all the algorithms on the MNIST handwritten digit dataset with $5\%$ outliers: Average recognition accuracy (AC) $\pm$ standard derivation.}
 \centering
 \scalebox{0.75}{
 \begin{tabular}{|p{60pt}<{\centering}| p{95pt}<{\centering} | p{95pt}<{\centering}| p{95pt}<{\centering} | p{95pt}<{\centering}|p{95pt}<{\centering}|}
   \hline
   \multirow{2}{*}{Methods}&
   \multicolumn{4}{c|}{Images per digit $\times$ $\sharp$ of digits}\\
   \cline{2-5}
  & $200\times10$ & $400\times10$& $600\times10$ & $800\times10$\\
   \hline
   2DPCA & 0.5643 $\pm$ 0.0300 & 0.6264 $\pm$ 0.0206& 0.6543 $\pm$ 0.0171 & 0.6788 $\pm$ 0.0136  \\
   \hline
   $L_1$-2DPCA & 0.5636 $\pm$ 0.0303 & 0.6257 $\pm$ 0.0204 & 0.6539 $\pm$ 0.0171 & 0.6782 $\pm$ 0.0136   \\
   \hline
   F-2DPCA  & 0.5759 $\pm$ 0.0216 & 0.6272 $\pm$ 0.0164 & 0.6490 $\pm$ 0.0119 & 0.6759 $\pm$ 0.0122  \\
   \hline
   2DSVD & 0.5865 $\pm$ 0.0215 & 0.6360 $\pm$ 0.0160 & 0.6565 $\pm$ 0.0121 & 0.6840 $\pm$ 0.0113  \\
   \hline
  $R_1$-2DSVD & 0.5860 $\pm$ 0.0212 & 0.6358 $\pm$ 0.0162 & 0.6562 $\pm$ 0.0121 & 0.6562 $\pm$ 0.0121  \\
   \hline
   N-2DNPP & 0.5925 $\pm$ 0.0223 & 0.6405 $\pm$ 0.0130 & 0.6548 $\pm$ 0.0160 & 0.6689 $\pm$ 0.0131  \\
   \hline
  S-2DNPP & 0.5675 $\pm$ 0.0304 & 0.6283 $\pm$ 0.0213 & 0.6566 $\pm$ 0.0154 & 0.6799 $\pm$ 0.0136  \\
   \hline
   Proposed  & \textbf{0.8326 $\pm$ 0.0022} & \textbf{0.8462 $\pm$ 0.0041}& \textbf{0.8458 $\pm$ 0.0014} & \textbf{0.8639 $\pm$ 0.0020}  \\
   \hline
 \end{tabular}}
 \end{table*}

\begin{figure}[H]
\centering
    \hspace{0cm}
    \vspace{0cm}
   \subfigure[]{\includegraphics[width=0.7\columnwidth]{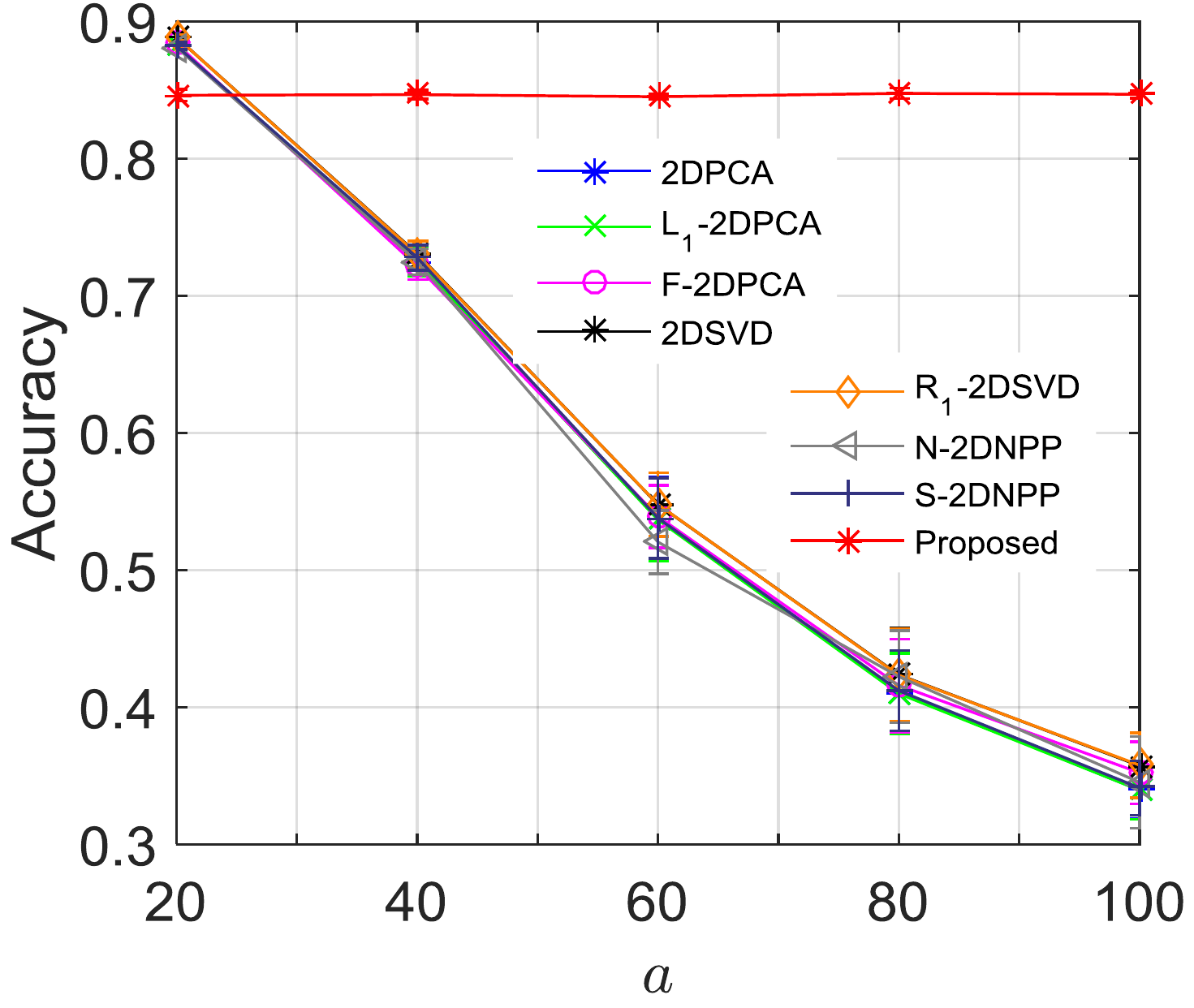}}
   \hspace{0cm}
   \subfigure[]{\includegraphics[width=0.7\columnwidth]{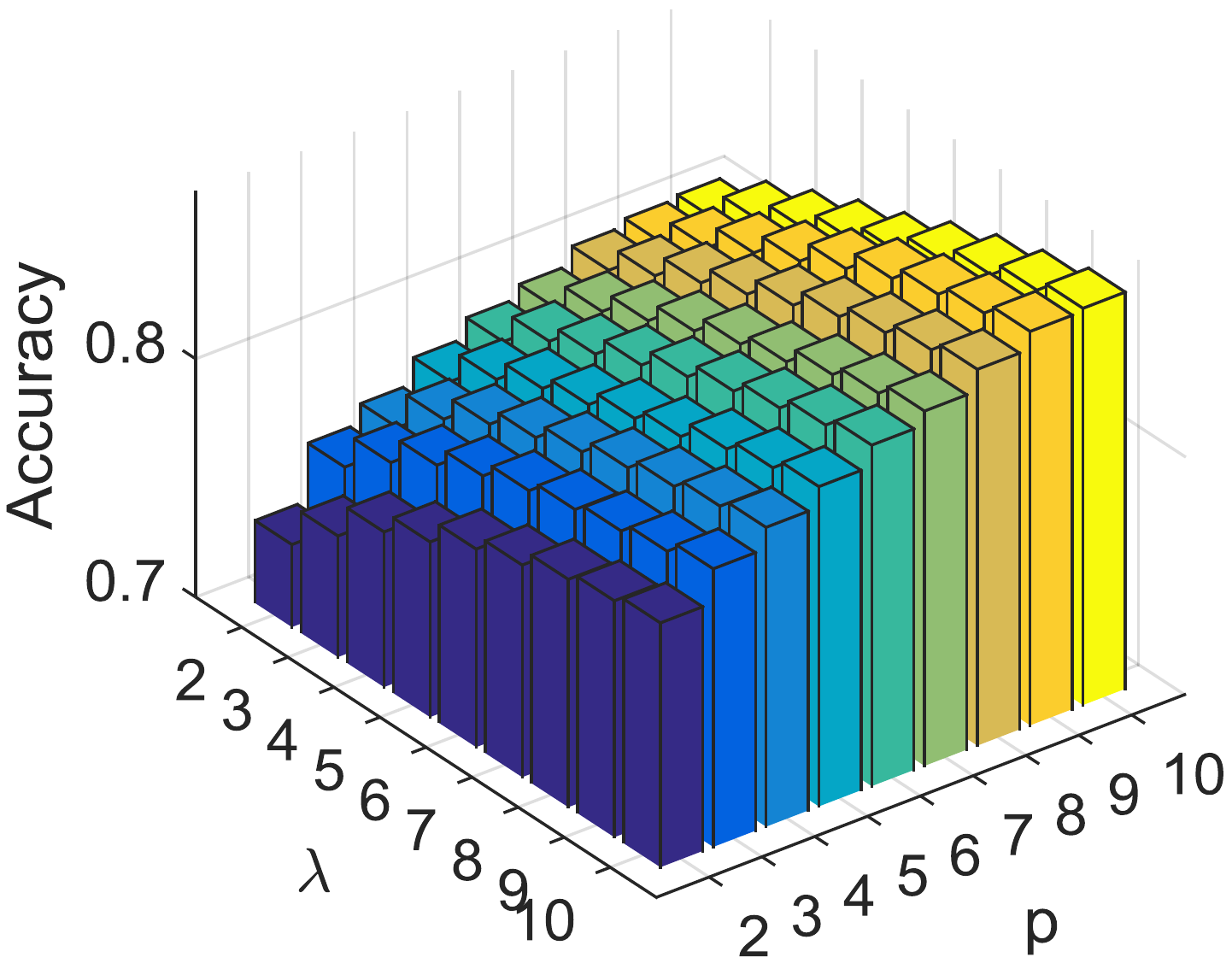}}
    \caption{Recognition accuracies on the MNIST Handwritten Digit Database. (a) Recognition accuracies of all the algorithms with changing magnitude of outliers ; (b) Recognition accuracies of the proposed algorithm with different $\lambda$ and $p$.}
\end{figure}

\begin{table}
 \caption {K-means clustering results of subspaces learned from different algorithms on the first 100 faces of the ORL database: Average Clustering Accuracy (AC) $\pm$ Standard Deviation and Average normalized mutual information (NMI) $\pm$ Standard Deviation.}
 \centering
 \scalebox{0.75}{
 \begin{tabular}{|p{60pt}<{\centering}|p{25pt}<{\centering}| p{93pt}<{\centering}| p{93pt}<{\centering} | p{93pt}<{\centering} |p{93pt}<{\centering}|}
    \hline
    \multicolumn{2}{|l|}{ \multirow{2}{*}{Methods}}&
   \multicolumn{4}{c|}{Number of principal components }\\
   \cline{3-6}
   \multicolumn{2}{|l|}{ } & $m=30$ & $m=50$ & $m=70$ & $m=90$ \\
   \hline
   \multirow{2}*{2DPCA}& AC & 0.5991 $\pm$ 0.0442 & 0.7535 $\pm$ 0.0153 & 0.8143 $\pm$ 0.0190 & 0.7507 $\pm$ 0.0070 \\
    & NMI & 0.7619 $\pm$ 0.0268 & 0.8692 $\pm$ 0.0042 & 0.8860 $\pm$ 0.0052 & 0.8684 $\pm$ 0.0019  \\
   \hline
   \multirow{2}*{$L_1$-2DPCA}& AC & 0.6981 $\pm$ 0.0176 & 0.8199 $\pm$ 1.2e-15 & 0.8003 $\pm$ 0.0315 & 0.7500 $\pm$ 0 \\
    & NMI & 0.8221 $\pm$ 0.0112 & 0.8875 $\pm$  1.4e-15 & 0.8821 $\pm$ 0.0087 & 0.8682 $\pm$ 4.4e-16  \\
   \hline
   \multirow{2}*{F-2DPCA} & AC & 0.7000 $\pm$ 1.3e-15  & 0.8199 $\pm$ 1.2e-15 & 0.7528 $\pm$ 0.0137 & 0.7500 $\pm$ 0 \\
    & NMI & 0.8200 $\pm$ 7.8e-16 & 0.8875 $\pm$ 1.4e-15 & 0.8690 $\pm$ 0.0038 & 0.8682 $\pm$ 4.4e-16 \\
    \hline
   \multirow{2}*{2DSVD}& AC &0.7012 $\pm$ 0.0836 & 0.7571 $\pm$ 0.0219 & 0.8108 $\pm$ 0.0236 & 0.7528 $\pm$ 0.0137\\
        & NMI & 0.8197 $\pm$ 0.0417 & 0.8615 $\pm$ 0.0136 & 0.8850 $\pm$ 0.0065 & 0.8690 $\pm$ 0.0038  \\
    \hline
   \multirow{2}*{$R_1$-2DSVD} & AC &
   0.6876 $\pm$ 0.0781 &  0.7615 $\pm$ 0.0165 & 0.8052 $\pm$ 0.0286&  0.7507 $\pm$ 0.0070  \\
    & NMI  &  0.8128 $\pm$ 0.0406 & 0.8640 $\pm$ 0.0095 & 0.8835 $\pm$ 0.0079 &  0.8684 $\pm$ 0.0019  \\
    \hline
   \multirow{2}*{N-2DNPP}& AC  &0.7975 $ \pm$ 0.0925 & 0.7822 $\pm$ 0.0351 & 0.7948 $\pm$ 0.0338 & 0.7528 $\pm$ 0.0138 \\
        & NMI  &  0.8753 $\pm$ 0.0295 & 0.8772 $\pm$ 0.0097 & 0.8806 $\pm$ 0.0093 & 0.8691 $\pm$ 0.0038 \\
   \hline
   \multirow{2}*{S-2DNPP}& AC &0.7411 $\pm$ 0.0250 & 0.7424 $\pm$ 0.0129 & 0.8163 $\pm$ 0.0177 & 0.7491 $\pm$ 0.0090 \\
        & NMI & 0.8223 $\pm$ 0.0148 & 0.8457 $\pm$ 0.0070 & 0.8859 $\pm$ 0.0082 & 0.8666 $\pm$ 0.0065  \\
   \hline
   \multirow{2}*{Proposed}& AC &\textbf{0.9160 $\pm$ 0.0479} & \textbf{0.9377 $\pm$ 0.0363} & \textbf{0.8461 $\pm$ 0.0721} & \textbf{0.7623 $\pm$ 0.0258} \\
        & NMI & \textbf{0.9158 $\pm$ 0.0241}  & \textbf{0.9292 $\pm$ 0.0191} & \textbf{0.8902 $\pm$ 0.0332} & \textbf{0.8704 $\pm$  0.0078} \\
   \hline
 \end{tabular}}
 \end{table}
\subsection{Experiments for Image Clustering}
 It has been proven in theory that dimensional reduction techniques can be used to improve the clustering accuracy as a preprocessing step~\cite{he2011robust, huang2008simultaneous, ding2004k}.  In this experiment, we test the proposed algorithm and all the benchmarks on a image clustering problem on the ORL face database in the presence of outliers. All the face images in the first 10 subjects are selectd to construct the training samples, and thus 100 images are selected as the training data. We random generate 30 dummy images as outliers and add them to the training data, thus the number of clean training samples and outlier samples are 100 and 30, respectively. For each algorithm, we apply Algorithm 1 on the training dataset, then K-means algorithm is applied to evaluate the quality of these features. Before applying the K-means, we initialize the clustering center by the density search based method proposed by Rodriguez and Laio~\cite{rodriguez2014clustering}.

To apply K-means algorithm, for the one-sided transforms, including 2DPCAs and 2DLPPs, we directly applied K-means on the projected samples (dimensional reduced), i.e., $X_i^{\text{new}}=\hat{X}_iU$, where $U$ is the projection matrix and $i=1,2,\cdots, N$. For the two-sided transforms including 2DSVDs and the proposed algorithm apply the K-means to the $\{M_i\}_{i=1}^N$. Two metrics are used to evaluate the clustering performance of each algorithm, one is the average clustering accuracy (AC) and another is the average normalized mutual information (NMI)~\cite{cai2005document}. The clustering results of different algorithm under varying number of principal components are shown in Table 2. To reduce the standard deviations, all the results are reported over 100 iterations. From the results in Table 2, we can see that the performance of the proposed algorithm is the best under different parameter settings.

To explore the effect of different $\lambda$ and $p$ values on the clustering performance of the proposed algorithm, we plot average clustering accuracy and normalized mutual information with varying $\lambda$ and $p$ in Figure 2. We can see that, better AC and NMI can be obtained by choosing a $p$ larger than 2. When $p$ is fixed, the clustering AC and NMI increase with $\lambda$ increasing.

 \begin{figure}[H]
\centering
    \hspace{0cm}
    \vspace{0cm}
   \subfigure[]{\includegraphics[width=0.7\columnwidth]{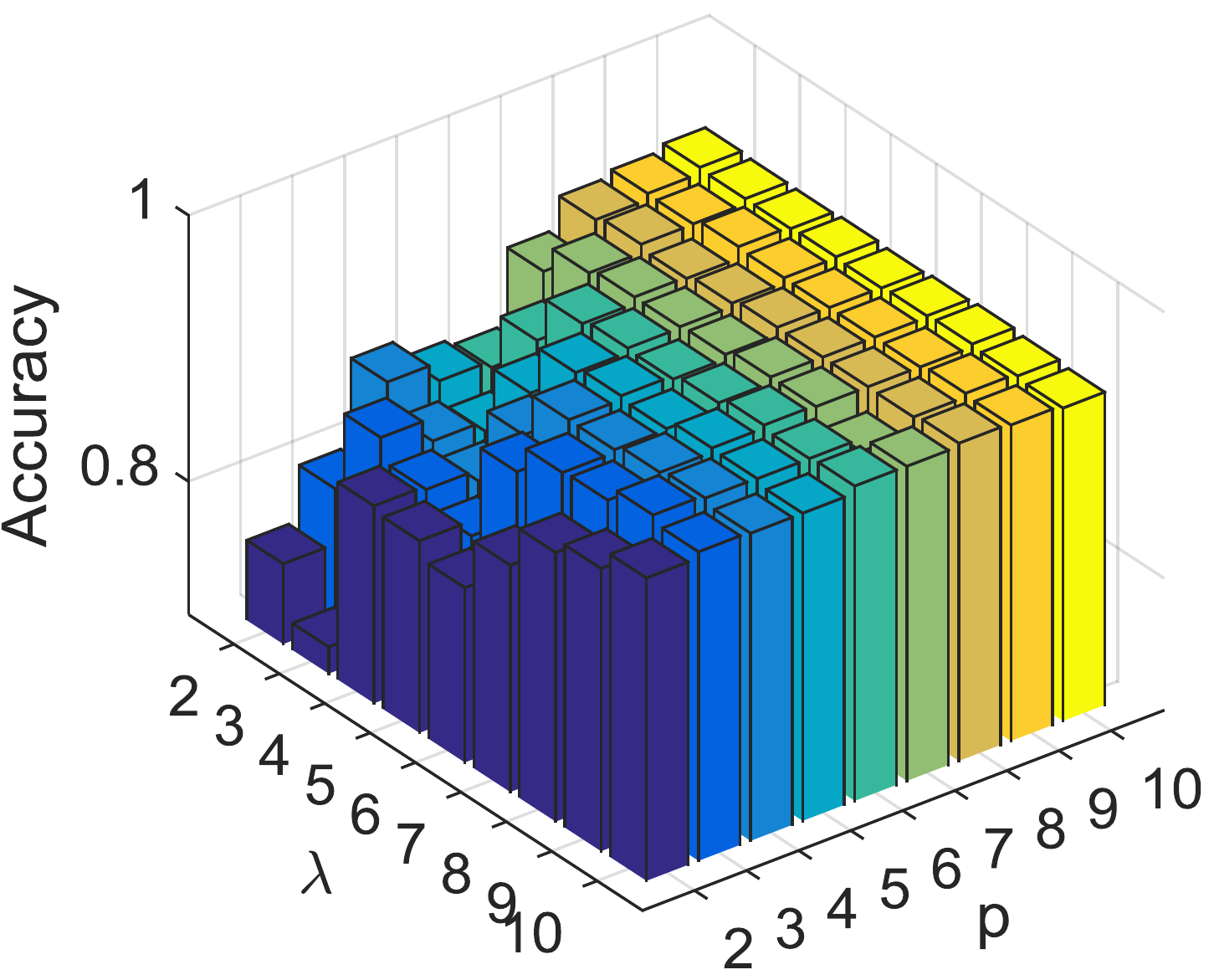}}
   \hspace{0cm}
   \subfigure[]{\includegraphics[width=0.7\columnwidth]{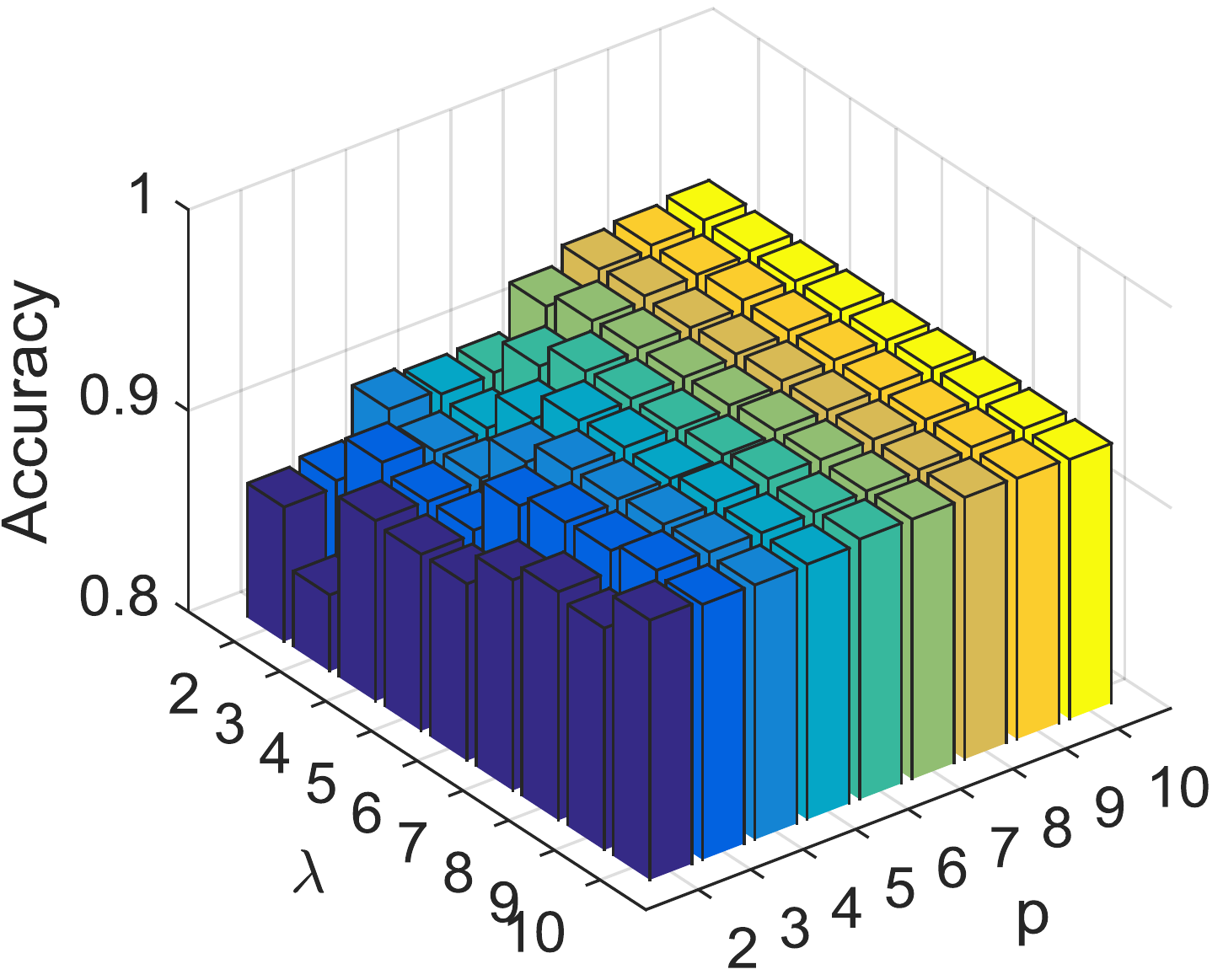}}
    \caption{Clustering performance with different $\alpha$ and $p$ on the ORL database. (a) Clustering accuracy; (b) clustering NMI. }
\end{figure}

 \subsection{Experiments for Image Reconstruction}
We then check the effectiveness of the proposed algorithm for image reconstruction on the Yale Face database. There are 165 face images in Yale Face database, we choose all the images as the training sample. To simulate outliers, we randomly generate 30 dummy image as in \cite{he2011maximum}. So the total number of training sample is 195 with 165 clean samples and 30 outlier samples. The reconstruction error is calculated by $\frac{1}{N}\sum_{i=1}^N \|{X}_i^{\text{org}}-X_i^{\text{new}}\|_2^2$, where ${X}_i^{\text{org}}$ and $X_i^{\text{new}}$ are the original training image and reconstructed image in the presence of outliers. For calculating the reconstruction error, the reconstruction for outlier images is excluded as in \cite{he2011maximum}, thus we just need calculate reconstruction errors for the 165 clean images. The average reconstruction errors of different algorithms are shown in Figure 3 (a) which shows that the proposed algorithm obtains the lowest reconstruction error after the number of principal is greater than 40. Images reconstructed by more principal components implies that more outlier information will be involved in reconstruction. Thus the results in Figure 3(a) show that the proposed algorithm has better ability in suppressing outlier information for image reconstruction. We also plot the average reconstruction errors under different parameter $\lambda$ and $p$ in Figure 3(b) to find out how does varying $\lambda$ and $p$ affect the image reconstruction performance of the proposed algorithm. We can see that a non-second $p$ order ($p$ is smaller than 2) offers better reconstruction results, and by choosing an appropriate $\lambda$ value, we can obtain the optimal results.
\begin{figure}[H]
\centering
    \hspace{0cm}
    \vspace{0cm}
   \subfigure[]{\includegraphics[width=0.7\columnwidth]{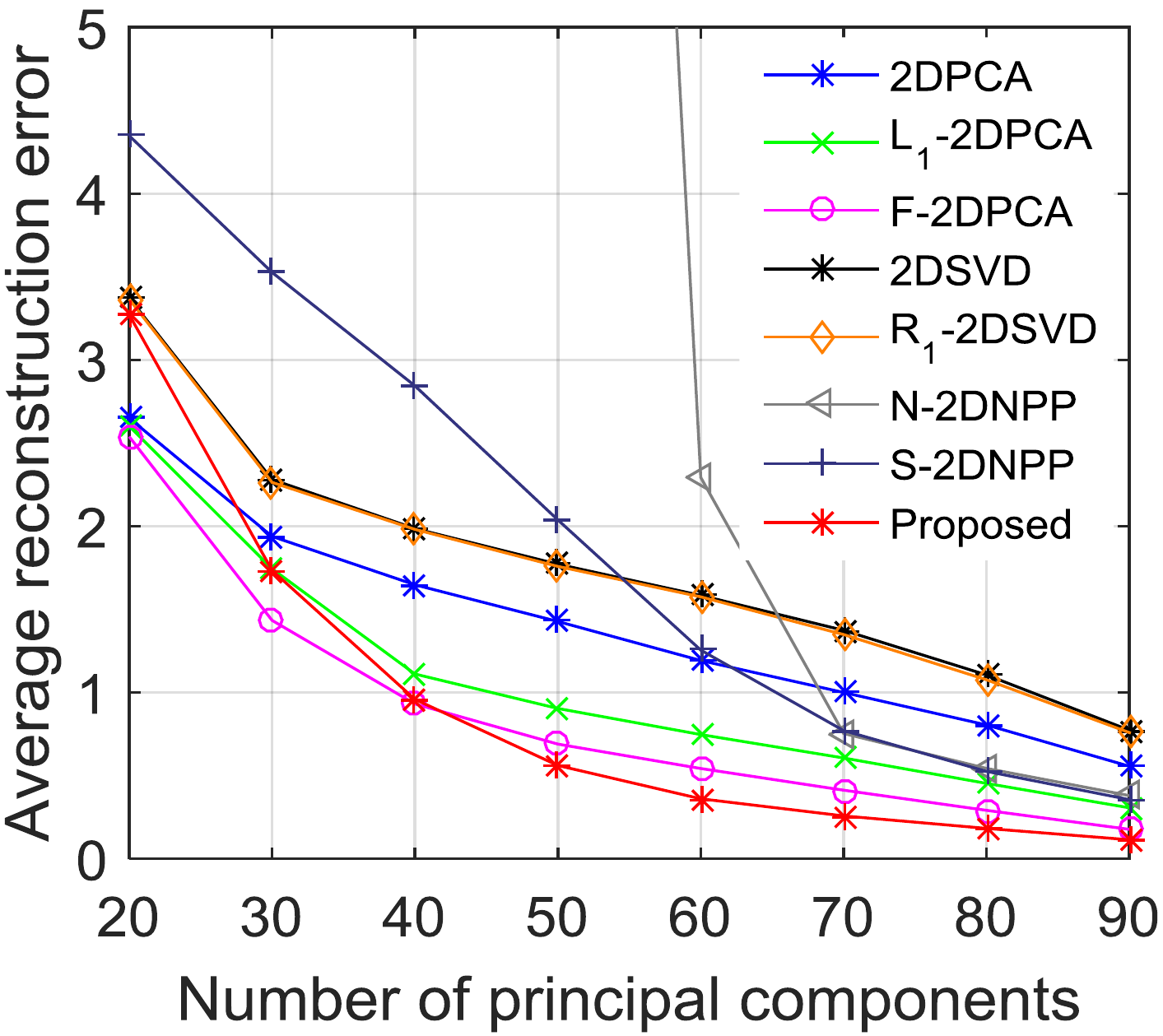}}
   \hspace{0cm}
   \subfigure[]{\includegraphics[width=0.7\columnwidth]{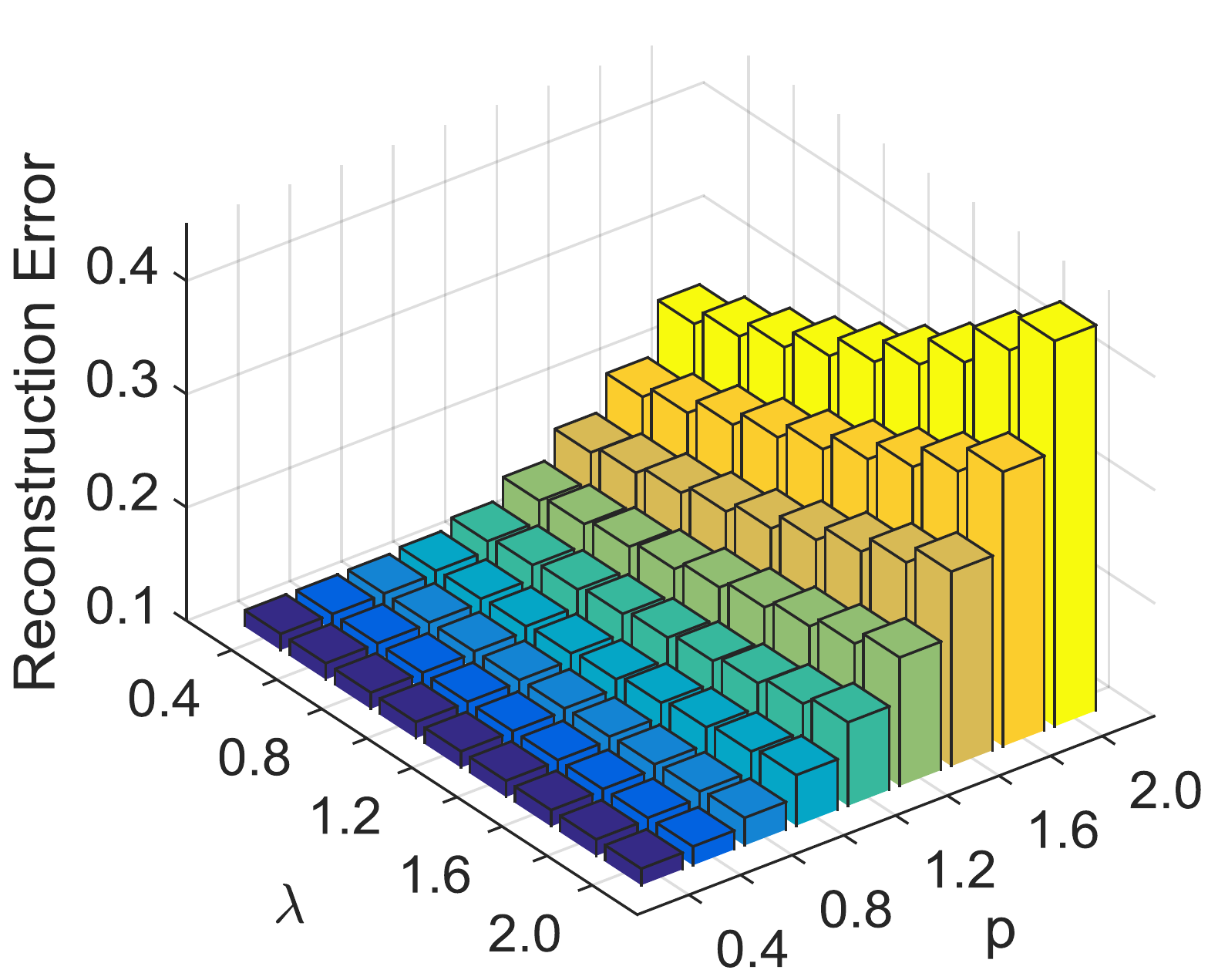}}
    \caption{Image reconstruction performance with different $\alpha$ and $p$ on the the Yale Face database. (a) Average reconstruction accuracy; (b) Reconstruction error with changing $\lambda$ and $p$. }
\end{figure}
 \section{Conclusion}
 In this paper, in order to better solve the outlier problem in the 2DSVD based algorithms, we developed a generalized kernel risk sensitive loss (GKRSL) for robust 2DSVD. Unlike the other 2DPCA and 2DSVD algorithms which treat each training sample equally, the GKRSL-2DSVD discriminatively weight the training samples so that the information of the outliers is excluded from the training procedure. Thus the learned features from the proposed model is more robust to outliers.  Since the resulted objective function is a non-convex, we developed a optimization algorithm based on the majorization minimization theory. A convergence analysis for the proposed objective function is also provided. Extensive experiments on three image processing applications on three public datasets with varying parameter settings show that the proposed algorithm has superior outlier-resistance ability to other benchmarks.


\section*{Acknowledgments}
This work is supported in part by Australian Research Council (ARC) under Discovery Grants DP140101075.


 \bibliographystyle{elsarticle-num}





\end{document}